\documentclass[10pt,twocolumn,letterpaper]{article}

\usepackage[pagenumbers]{cvpr} 

\usepackage{booktabs}
\usepackage{multirow}
\usepackage{graphicx}

\definecolor{cvprblue}{rgb}{0.21,0.49,0.74}
\usepackage[pagebackref,breaklinks,colorlinks,allcolors=cvprblue]{hyperref}

\def\paperID{12414} 
\def\confName{CVPR}
\def\confYear{2025}

\def\ours{MVideo}
\title{\ours: Motion Control for Enhanced Complex Action Video Generation}

\author{Qiang Zhou$^{1}$, Shaofeng Zhang$^{2}$, Nianzu Yang$^{2}$, Ye Qian$^{1}$, Hao Li$^{3}$\thanks{: Corresponding author}\\
$^{1}$INF Tech. $^{2}$Shanghai Jiao Tong University, $^{3}$Fudan University\\
{\tt\small 
\{zhouqiang, qianye.0514\}@inftech.ai, \{sherrylone, yangnianzu\}@sjtu.edu.cn, lihao\_lh@fudan.edu.cn}
}

\begin{document}
\maketitle
\begin{figure}[tb!]
    \centering
    \includegraphics[width=0.9\linewidth]{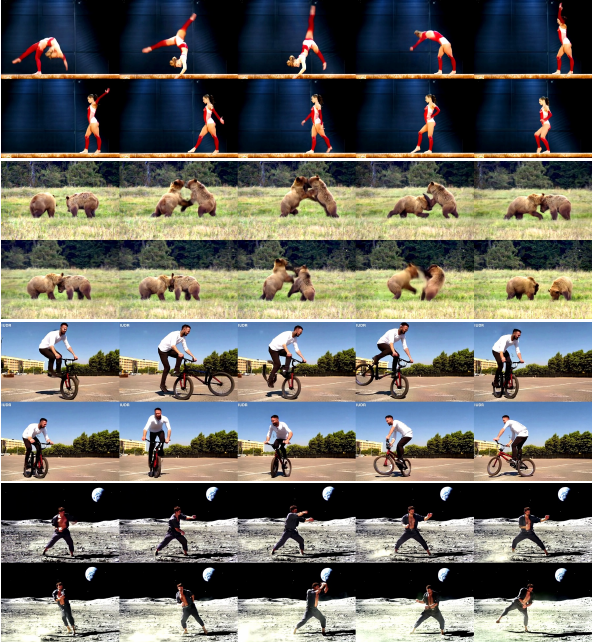}
    \caption{Complex action videos generated by MVideo, with a duration of 12 seconds and a spatial resolution of 480 $\times$ 720 pixels.}
    \label{fig:enter-label}
\end{figure}
\begin{abstract}
Existing text-to-video (T2V) models often struggle with generating videos with sufficiently pronounced or complex actions. A key limitation lies in the text prompt’s inability to precisely convey intricate motion details. To address this, we propose a novel framework, MVideo, designed to produce long-duration videos with precise, fluid actions. MVideo overcomes the limitations of text prompts by incorporating mask sequences as an additional motion condition input, providing a clearer, more accurate representation of intended actions. Leveraging foundational vision models such as GroundingDINO and SAM2, MVideo automatically generates mask sequences, enhancing both efficiency and robustness.
Our results demonstrate that, after training, MVideo effectively aligns text prompts with motion conditions to produce videos that simultaneously meet both criteria. This dual control mechanism allows for more dynamic video generation by enabling alterations to either the text prompt or motion condition independently, or both in tandem. Furthermore, MVideo supports motion condition editing and composition, facilitating the generation of videos with more complex actions.
MVideo thus advances T2V motion generation, setting a strong benchmark for improved action depiction in current video diffusion models.
Our project page is available at \href{https://mvideo-v1.github.io/}{https://mvideo-v1.github.io/}.
\end{abstract}    

\section{Introduction}
\label{sec:intro}

Despite significant progress in the field of image~\cite{esser2024scaling, zhang2024continuous, chen2023pixart,parihar2024precisecontrolenhancingtexttoimagediffusion,liang2024richhumanfeedbacktexttoimage,li2024autoregressiveimagegenerationvector,kolors,esser2024scalingrectifiedflowtransformers} and video generation~\cite{ho2022videodiffusionmodelsVDM,wang2023modelscopetexttovideotechnicalreport,ma2024lattelatentdiffusiontransformer,chen2024videocrafter2overcomingdatalimitations,wei2023dreamvideocomposingdreamvideos,weng2023artboldsymbolcdotvautoregressivetexttovideogeneration,wang2024magicvideov2multistagehighaestheticvideo,bartal2024lumierespacetimediffusionmodel,videoworldsimulators2024SORA,shi2024poseguidedfinegrainedsignlanguage,opensora,yang2024cogvideox,tian2024videotetriscompositionaltexttovideogeneration,zhou2024allegroopenblackbox,ma2024vidpanosgenerativepanoramicvideos,wei2024dreamvideo2zeroshotsubjectdrivenvideo,polyak2024moviegencastmedia,wang2024loonggeneratingminutelevellong,wang2024emu3nexttokenpredictionneed}, current models still face substantial challenges when tasked with generating complex action videos. One of the core difficulties lies in the inability of text-based descriptions to fully capture the nuanced details of intricate movements and actions. This limitation significantly hinders the model’s ability to accurately train and infer complex action sequences in videos. Furthermore, these action videos often span longer durations, increasing the complexity of the generation process by requiring models to maintain temporal coherence across extended periods.

In this work, we introduce MVideo, a novel framework specifically designed to tackle the inherent challenges of generating complex action videos. Traditional video generation models rely heavily on textual descriptions, which are often inadequate for conveying the dynamic intricacies of complex actions. To overcome this limitation, MVideo leverages mask sequences as an additional conditioning input. Unlike text prompts, mask sequences offer a more precise and explicit representation of the desired actions, allowing the model to generate videos that more accurately capture the intended movements. Thanks to advancements in foundational vision models, such as GroundingDINO~\cite{liu2024groundingdinomarryingdino} and SAM2~\cite{ravi2024sam2segmentimages}, the mask sequences can be extracted automatically, enhancing MVideo's efficiency and robustness in action video generation.
%
%
Existing video diffusion models can only generate a few seconds of video, which is insufficient for creating a complete and coherent motion sequence.
To address this, MVideo proposes an efficient iterative video generation method that combines image conditions with low-resolution video conditions. This method reduces the computational cost of the model while maintaining temporal consistency, ensuring that even longer videos maintain coherent content and consistent action sequences throughout their duration.
MVideo is fine-tuned from the existing video diffusion model CogvideoX~\cite{yang2024cogvideox}. However, through experimentation, we observed that simply adding the mask sequence condition during fine-tuning led to a noticeable decline in the model’s ability to align text prompts with video generation, as reflected in the reduced metrics for ``overall consistency'' and ``imaging quality'' shown in Table~\ref{tbl:ablation_consistency_loss}. To mitigate this issue, we propose a novel consistency loss during training, which distills the text-to-video generation expertise of CogvideoX into MVideo, ensuring that the model not only learns to align mask sequences but also preserves its original text-condition capabilities.

Our experiments demonstrate that MVideo effectively aligns text prompts with motion conditions and generalizes to previously unseen mask sequences. This deep alignment capability enables MVideo to generate complex video effects. For instance, given a motion condition, MVideo allows modification of foreground objects or background scenes via text prompts and supports editing or combining motion conditions to produce more intricate actions. Additional examples are provided in the case study section~\ref{sec:case_study}.
In summary, our contributions of MVideo are threefold:


\begin{itemize}
    \item We introduce MVideo, a novel framework that iteratively generates long-duration action videos by integrating additional motion conditions for precise motion control.

    \item We show that MVideo generalizes effectively, aligning with unseen motion conditions and enabling complex video generation through motion condition editing and combination.

    \item  We validate MVideo’s effectiveness through quantitative and visual comparisons with state-of-the-art video diffusion methods.
\end{itemize}

\section{Related Work}
\label{sec:relate}

\paragraph{Video diffusion model.}

Video diffusion models (VDMs)~\cite{ho2022videodiffusionmodelsVDM} have pioneered video generation by extending traditional image diffusion U-Net~\cite{ronneberger2015UNet} architectures into 3D U-Net structures, employing joint training with image and video data to address video generation tasks effectively.
In contrast to VDMs, Make-A-Video~\cite{singer2022makeavideotexttovideogenerationtextvideo} innovates by learning visual-text correlations from paired image-text data and capturing motion dynamics from unsupervised video data, enabling flexible, scalable generation. 
MagicVideo~\cite{zhou2023magicvideoefficientvideogeneration} and LVDM~\cite{he2023latentvideodiffusionmodels} extend VDMs by implementing the Latent Diffusion Model (LDM) for text-to-video (T2V) generation, with LVDM employing a hierarchical framework to enhance latent representation modeling.
VideoFactory~\cite{wang2024swapattentionspatiotemporaldiffusions} introduces a swapped cross-attention mechanism, optimizing temporal and spatial interactions, while ModelScope~\cite{wang2023modelscopetexttovideotechnicalreport} advances T2V with spatial-temporal convolution and attention in LDM, enriching feature representation and motion comprehension.
Show-1~\cite{zhang2023show1marryingpixellatent} combines pixel- and latent-based diffusion models, showcasing the benefits of integrating different paradigms. SVD~\cite{blattmann2023stablevideodiffusionscalingSVD} further validates VDMs' adaptability across data scales and complexities.
Recently, CogvideoX~\cite{yang2024cogvideox} set a new standard for long-duration, coherent video generation with significant motion dynamics. 

Our model, MVideo, builds on CogvideoX~\cite{yang2024cogvideox}, finetuning it to enhance action generation capabilities, advancing video diffusion model performance.

\paragraph{Motion control.}

MCDiff~\cite{chen2023motionconditioneddiffusionmodelcontrollableMCDiff} is the first approach to use motion as a condition for controlled video synthesis. It takes the first frame of a video and a sequence of motion strokes as input.
DragNUWA~\cite{yin2023dragnuwafinegrainedcontrolvideo} combines text, image, and trajectory inputs for precise control over video content in semantic, spatial, and temporal dimensions. To address limitations in open-domain trajectory control, it introduces three innovations: a Trajectory Sampler (TS) for arbitrary trajectory control, Multiscale Fusion (MF) for granular trajectory control, and an Adaptive Training (AT) strategy to ensure trajectory-consistent video generation.
MotionCtrl~\cite{wang2024motionctrlunifiedflexiblemotion} presents two control modules for camera and object motion. It proposes specific dataset collection methods for these motion types, allowing the trained modules to be integrated into various video diffusion models for enhanced motion control.
Direct-a-Video~\cite{Yang_2024_direct_a_video} propose a strategy for decoupling control of object and camera motion, while MotionBooth~\cite{wu2024motionbooth} animates customized subjects with precise control over both object and camera movements.
%
FreeTraj~\cite{qiu2024freetraj} introduces a tuning-free
framework to achieve trajectory-controllable video generation, by imposing guidance on both noise construction and attention computation. 
VMC~\cite{jeong2023vmc} presents a novel one-shot tuning approach, focusing on adapting temporal attention layers within video diffusion models to generate motion-driven videos.
%
Motion Inversion~\cite{wang2024motioninversion} introduces a simple yet effective motion embedding that is disentangled
from appearance and suitable for motion customization tasks
MotionDirector~\cite{zhao2023motiondirector} propose a dual-path architecture and a novel appearance-debiased temporal training objective, to decouple the learning of appearance and motion

Previous works use trajectory lines to control target motion, but they struggle with complex actions, such as backflips. In contrast, our approach, MVideo, employs more precise mask sequences to generate videos of complex actions effectively.


\section{Framework}
\label{sec:framework}

\begin{figure*}
    \centering
    \includegraphics[width=0.85\linewidth]{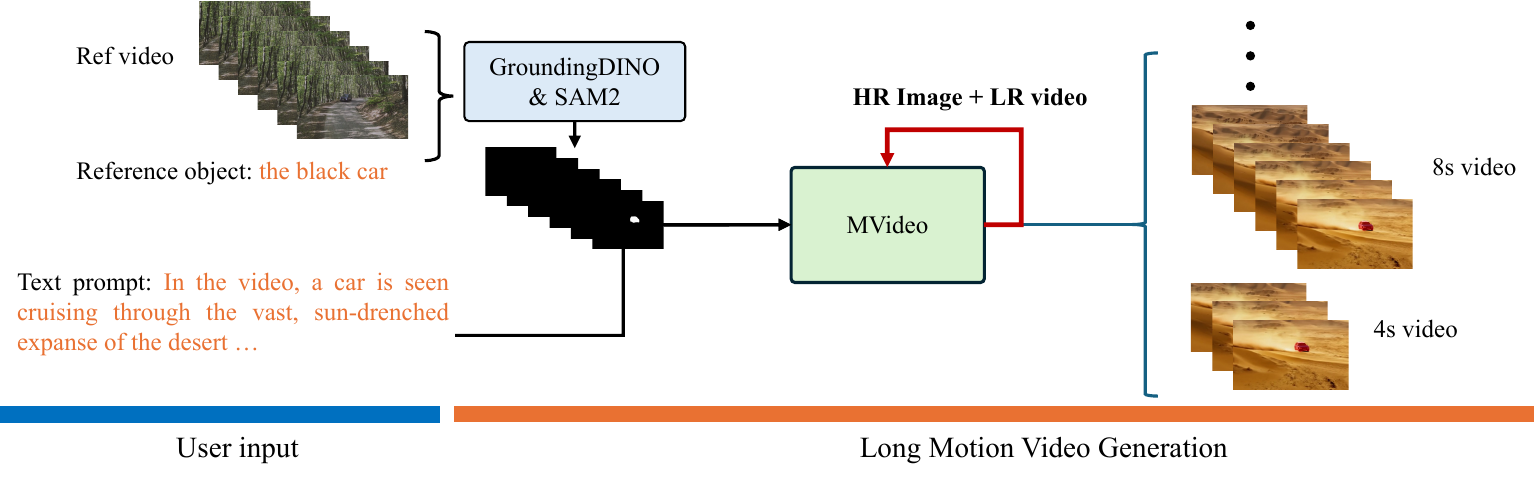}
    \caption{\textbf{Inference Pipeline:} Given a text prompt and a simple description identifying the reference object in the video, MVideo automatically extracts the corresponding mask sequence and iteratively generates long-duration motion videos, using the mask sequence as an additional condition.}
    \label{fig:pipeline_inference}
\end{figure*}

\begin{figure*}
    \centering
    \includegraphics[width=0.9\linewidth]{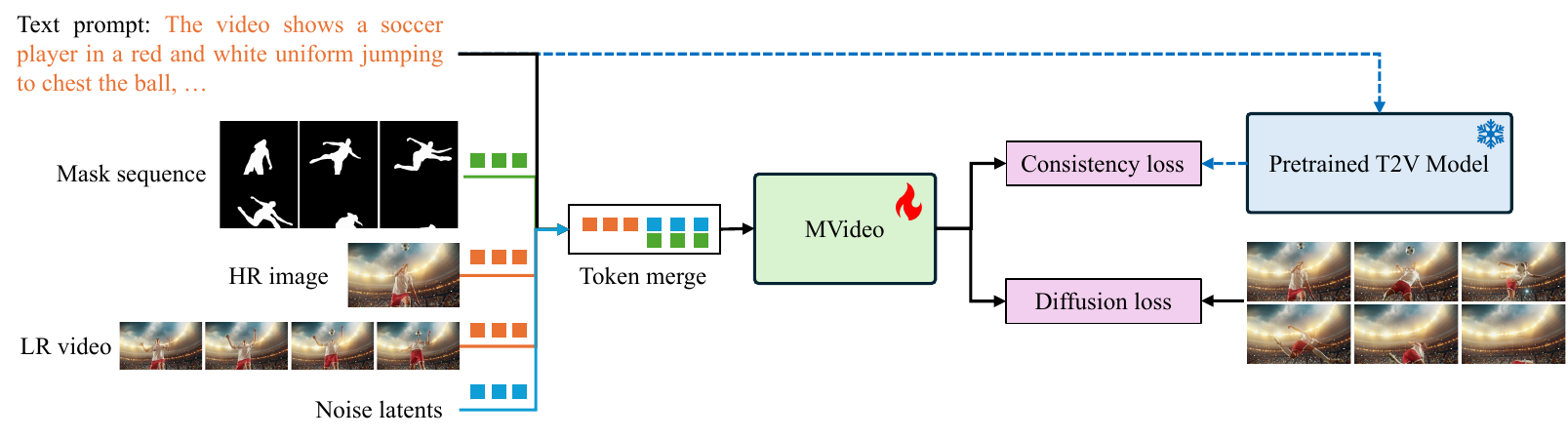}
    \caption{\textbf{Training Pipeline:} MVideo takes a text prompt as input, along with a mask sequence, a high-resolution image, and a low-resolution video as conditions. During training, diffusion loss adapts these new conditions, while consistency loss preserves the model's text-alignment capability.}
    \label{fig:pipeline_train}
\end{figure*}

In this section, we introduce the MVideo framework in detail, covering the motion conditions, long-duration motion video generation, and training objectives.

\subsection{Preliminary}
\label{sec:preliminary}

Video diffusion models are a type of probabilistic generative
model that simulates a forward and reverse diffusion process. The forward process involves gradually adding noise
to video data, transforming it into a standard Gaussian distribution.
The reverse process, which the model learns, involves denoising and reconstructing the original video data from the noise.
The training objective is to learn the reverse diffusion process, typically through a loss function that measures the
discrepancy between the predicted noise and the true noise
added at each step. The commonly used loss function in video diffusion models is the Mean Squared Error (MSE) between
the true noise $\epsilon$ and the predicted noise $\hat{\epsilon}$. The loss formula is:
\begin{equation}
    \mathcal{L} = \mathbb{E}_{\mathbf{x}, \epsilon, \mathbf{c}, t}[|| \epsilon - \hat{\epsilon}(\textbf{x}_t, \mathbf{c}, t) ||_{2}^2]
\end{equation}

where $\mathbf{x}$ is the original video data, $\mathbf{x}_t$ is the noisy data at step $t$, $\mathbf{c}$ is the text prompt,
$\epsilon$ is the true noise added to $\mathbf{x}$, and $\hat{\epsilon}(\mathbf{x}_t, \mathbf{c}, t)$ is the model’s predicted noise.
Moreover, rather than reconstructing the raw video data directly, a video VAE~\cite{kingma2022autoencodingvariationalbayesVAE} encoder is typically used to first transform the raw video into compressed latent representations. This method notably decreases the computational demands on the video diffusion model.

\subsection{Motion Control}
\label{sec:object_motion}

\paragraph{Mask-to-Video.}
Current video diffusion models often struggle to depict intricate object movements accurately, as text prompts alone are frequently insufficient to specify complex action details. For generating high-complexity action videos, more precise guidance on object actions is essential. In this work, we demonstrate that mask sequences can effectively supplement or replace traditional text prompts. By integrating mask sequences with textual descriptions, our approach enhances the model's capability to produce videos with intricate object actions.

\paragraph{Mask Extraction.}
Efficient mask sequence extraction is crucial for the practical application of video diffusion models. Recent advances in foundational vision models provide robust support for this task, facilitating mask extraction across diverse objects.
In our method, we employ the GroundingDINO \cite{liu2024groundingdinomarryingdino} and SAM2 \cite{ravi2024sam2segmentimages} models to automate the generation of mask sequences, as illustrated in Figure~\ref{fig:pipeline_inference}. 
Given a reference video and a description of the target object, we first utilize the GroundingDINO model to identify the bounding box of the object in the initial frame. This bounding box is then input to the SAM2 model, which segments the mask sequence of the target object across the entire video. 
Aside from a basic text prompt, our framework requires only a reference video and an object description, enabling the automated extraction of the mask sequence as a conditioning input for motion.

\paragraph{Mask Fusion.}
Several approaches can be used to incorporate the mask sequence condition into the diffusion model. Based on our experiments, we found channel concatenation to be particularly effective. This method offers the advantage of not increasing the parameter count or computational complexity of the diffusion transformer, ensuring compatibility with various pretrained video diffusion models. Specifically, we first employ a motion encoder to extract motion features from the mask sequences, which are then fused with the noisy latent representations through channel-wise concatenation, as shown in Figure~\ref{fig:pipeline_train}. Our experiments show that a pretrained video VAE encoder performs effectively as the motion encoder, allowing us to use it without further optimization during training.

\subsection{Long Motion Video}
\label{sec:long_motion}

Generating complex action videos, such as Tai Chi performances, requires longer durations to accurately capture a sequence of intricate actions. However, existing pretrained video diffusion models are generally limited to producing clips only a few seconds long, which falls short for creating such complex action videos.

In this work, we propose an efficient approach for recursively generating high-quality, long-duration videos. Our method involves generating video clips of $t_1$ seconds each (defaulting to 4 seconds) and concatenating them to form the final video. To ensure temporal consistency in appearance and motion across the extended video, we introduce two additional conditioning mechanisms: a high-resolution appearance condition and a low-resolution motion condition, as illustrated in Figure~\ref{fig:pipeline_train}.
The appearance condition enhances visual consistency by incorporating high-resolution frames—specifically, the final frame from the preceding clip—to align the appearance of the current clip with previous ones. For motion consistency, we introduce a motion condition that maintains coherent motion throughout the video by using low-resolution video segments of $t_2$ seconds (also defaulting to 4 seconds) from prior clips as input. To balance efficiency with quality, these motion conditions are set to a low resolution of $256\times 384$.
Both the appearance and motion conditions are encoded as latent features using a video VAE encoder, and these features are concatenated with the noisy latent representations along the sequence dimension, as shown in Figure~\ref{fig:pipeline_train}.
Experimental results demonstrate that our framework effectively generates long-duration action videos while maintaining strong temporal consistency in both content and motion.

\subsection{Training Loss}

MVideo is finetuned from the pretrained video diffusion model, CogVideoX~\cite{yang2024cogvideox}, using a custom mask-to-video training dataset. This finetuning process enhances MVideo's mask sequence alignment performance but results in decreased text alignment ability and overall video quality. As shown in Table~\ref{tbl:ablation_consistency_loss}, the evaluation metrics for ``overall consistency" and ``imaging quality" on the VBench~\cite{vbench_HuangHYZS0Z0JCW24} test set have declined significantly.
To improve MVideo’s mask sequence alignment while retaining the text alignment capabilities learned by the original CogVideoX model, we introduced a consistency loss term in addition to the diffusion loss. This consistency loss is computed as the L2 loss between MVideo’s predictions and CogVideoX’s predictions, serving to maintain alignment between the two models' outputs. Our ablation studies confirm that incorporating this consistency loss effectively preserves MVideo’s text alignment capabilities without compromising mask sequence alignment learning.
The final loss function for MVideo training is therefore defined as follows:
\begin{equation}
    \mathcal{L} = \mathcal{L}_{d} + \alpha ~ \mathcal{L}_{c},
\end{equation}
where $\mathcal{L}_{d}$ is the diffusion loss, $\mathcal{L}_{c}$ is the consistency loss, and $\alpha$ represents the loss weight, set to 1.0 by default.

\section{Experiments}

\subsection{Experimental setup}

\paragraph{Implementation details.}

MVideo is initialized from CogVideoX~\cite{yang2024cogvideox}, a highly effective open-source video diffusion model. To conserve training resources, we freeze the parameters of CogVideoX-5b and employ LoRA fine-tuning to incorporate capabilities such as mask sequence conditioning and long video generation.
MVideo is trained using 32 GPUs with a total batch size of 64 over 10,000 steps. The learning rate is set to $2 \times 10^{-4}$, and the optimizer utilized is Adam. The default resolution for video generation is set to $480 \times 720$. 
For videos longer than 4 seconds, the resolution for the low-resolution video condition defaults to $256 \times 384$.

\paragraph{Training data.}

The training of \ours~ requires triplet data in the format of $<$video, text, mask sequence$>$. To prepare these training data, we automatically generated 400,000 samples using the GroundingDINO~\cite{liu2024groundingdinomarryingdino} and SAM2~\cite{ravi2024sam2segmentimages} models, which we refer to as the mask-to-video training set.
Specifically, we performed scene cuts on the original HDVILA~\cite{xue2022hdvila} dataset and then selected video clips longer than 8 seconds from the resulting segments. For each video clip, we initially employed the GroundingDINO model to identify bounding boxes around objects in the first frame. These bounding boxes were then used to initialize the SAM2 model, which subsequently extracted object masks across all frames of the video. Since extracting mask sequences for every object in a video is computationally intensive, we focused on the 80 object categories defined in the MSCOCO~\cite{LinMBHPRDZ14MSCOCO} dataset. Our subsequent experiments indicate that after training MVideo on these mask sequences, the model effectively generalizes to mask sequences of objects beyond these 80 categories during testing. 
Finally, we utilized the VILA~\cite{xue2022hdvila} model to extract detailed video captions for these video clips.

\paragraph{Testing data.}

\begin{table}[!t]
\centering
\resizebox{0.9\linewidth}{!}{%

\begin{tabular}{@{}lc|c|c|c@{}}
\toprule
\multicolumn{2}{c|}{Data}                                     & text & mask sequence & number of samples \\ \midrule
\multicolumn{1}{c|}{Trainset}                 & Mask-to-Video &   $\checkmark$   &    $\checkmark$            & 400,000        \\ \midrule

\multicolumn{1}{c|}{\multirow{3}{*}{Testset}} & VBench~\cite{vbench_HuangHYZS0Z0JCW24}        &  $\checkmark$     &               & 100            \\ 
\multicolumn{1}{c|}{}                         & Mask-to-Video &  $\checkmark$     &  $\checkmark$              & 500            \\
\multicolumn{1}{c|}{}                         & ComplexMotion &   $\checkmark$    &   $\checkmark$             & 200            \\ \bottomrule
\end{tabular}
}
\caption{Statistics of the training and testing data used in MVideo.}
\label{tbl:dataset}
\end{table}

We utilize multiple test sets to systematically evaluate the performance of MVideo, including the VBench~\cite{vbench_HuangHYZS0Z0JCW24} test set, the Mask-to-Video test set, and the ComplexMotion test set.
The VBench test set does not contain mask sequences and is primarily used to assess MVideo's text-to-video generation capabilities across various video durations.
The Mask-to-Video test set is collected from the HDVILA dataset and includes mask sequences that do not appear in the training data, allowing us to evaluate MVideo's generalization performance in mask alignment.
The ComplexMotion test set is specifically curated from the Pexels website to include videos featuring complex motion, enabling an assessment of MVideo's generation quality in challenging dynamic scenarios.

\paragraph{Metrics.}
For the quantitative evaluation metrics, we primarily employ the metrics provided by VBench~\cite{vbench_HuangHYZS0Z0JCW24} to assess various aspects of the generated videos, including overall consistency and motion smoothness. Additionally, we introduce a novel metric called mask mIoU, which specifically evaluates the alignment between the videos generated by MVideo and the corresponding mask conditions.
Specifically, for each generated video frame produced by \ours, we use the bounding box from the ground truth mask $m$ to initialize the SAM2 model, which then extracts the corresponding mask $\hat{m}$ from the generated frame. The mask Intersection over Union (IoU) is calculated between masks $m$ and $\hat{m}$. Finally, the mask mIoU metric $S_m$ is calculated as the average IoU across all videos and frames in the test set,
\begin{equation}
    S_m = \frac{1}{NM}\sum_{i,j}{\text{IoU}(m_{i,j}, ~ \hat{m}_{i,j})},
\end{equation}
where $N$ represents the number of generated videos, $M$ denotes the number of frames in each video, $i \in \{0, 1, \ldots, N-1\}$ is the index of each video, and $j \in \{0, 1, \dots, M-1\}$ is the index of the frames within a video.

\subsection{Results}

\begin{figure*}[!t]
    \centering
    \includegraphics[width=1.0\linewidth]{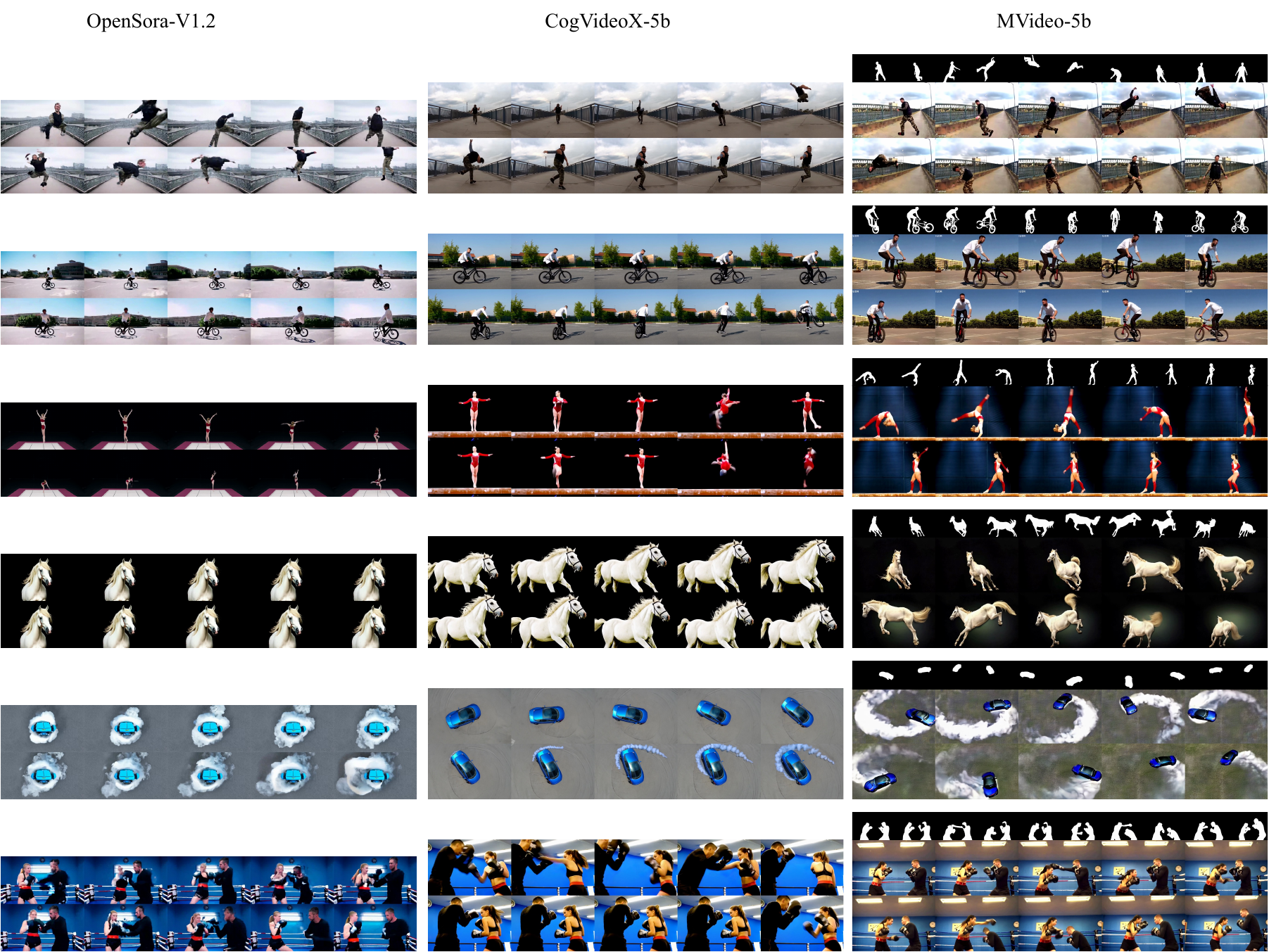}
    \caption{Visual comparison of videos generated by OpenSora-v1.2~\cite{opensora}, CogVideoX-5b~\cite{yang2024cogvideox}, and MVideo-5b.}
    \label{fig:vis_compare}
\end{figure*}

MVideo is conditioned on both text and mask sequences, and the alignment capabilities for these two conditions are evaluated separately. For text alignment, we use the VBench~\cite{vbench_HuangHYZS0Z0JCW24} test set, while for the newly introduced mask sequence alignment, we use the collected mask-to-video test set and the ComplexMotion test set.

\begin{table}[!t]
\centering
\resizebox{1.0\linewidth}{!}{%
\begin{tabular}{@{}c|c|ccc@{}}
\toprule
\multirow{2}{*}{Model} & \multirow{2}{*}{Video} & \multicolumn{3}{c}{VBench~\cite{vbench_HuangHYZS0Z0JCW24} testset}                                                                  \\ 
                       &                        & \multicolumn{1}{c|}{overall consistency} & \multicolumn{1}{c|}{imaging quality} & motion smoothness \\ \midrule
OpenSora-v1.2~\cite{opensora}          & $4s \times 720 \times 1280$        & \multicolumn{1}{c|}{25.63}               & \multicolumn{1}{c|}{62.76}           & 98.95             \\ 
CogvideoX-2b~\cite{yang2024cogvideox}           & $6s \times 480 \times 720$         & \multicolumn{1}{c|}{25.45}               & \multicolumn{1}{c|}{61.48}           & 98.02             \\ 
CogvideoX-5b~\cite{yang2024cogvideox}           & $6s \times 480 \times 720$         & \multicolumn{1}{c|}{26.29}               & \multicolumn{1}{c|}{61.55}           & 97.44             \\ \midrule
MVideo-5b              & $12s \times 480 \times 720 $        & \multicolumn{1}{c|}{26.57}               & \multicolumn{1}{c|}{60.53}           & 97.19             \\ 
\bottomrule
\end{tabular}
}
\caption{Text alignment performance comparison.}
\label{tbl:compare_text_align}
\end{table}

\paragraph{Text alignment.}

Table~\ref{tbl:compare_text_align} shows that on the VBench test set, MVideo-5b achieves comparable performance in overall consistency, image quality, and motion smoothness to OpenSora-v1.2 and CogVideoX-5b, indicating robust text-to-video generation capabilities. Ablation studies suggest that this performance is largely due to the consistency loss applied during training.

\begin{table}[!t]
\centering
\resizebox{0.9\linewidth}{!}{%
\begin{tabular}{@{}l|c|c|c@{}}
\toprule
Model     & Video           & Mask-to-Video testset & ComplexMotion testset \\ \midrule
MVideo-5b & $12s \times 480 \times 720$ &      77.90        &   78.34                    \\ \bottomrule
\end{tabular}
}
\caption{Mask sequence alignment performance evaluated using the mask mIoU metric $S_m$.}
\label{tbl:result_mask_miou}
\end{table}

\paragraph{Mask sequence alignment.}

The mask-to-video training set for MVideo was created using 80 object names from the MSCOCO dataset. To evaluate generalization, we tested MVideo’s alignment performance on unseen objects by constructing a mask-to-video test set with triplets in the form $<$video, text, mask sequence$>$. This test set includes mask sequences from objects absent in the training data, such as "cheetah," "deer," "llama," "penguin," "sea turtle," "squirrel," and "tiger," with all videos sourced from the public HDVILA dataset.
As shown in Table~\ref{tbl:result_mask_miou}, MVideo exhibits strong generalization in mask alignment, achieving high mIoU scores on unseen object mask sequences. Even with novel mask sequences, MVideo consistently aligns masks accurately in generated videos. Notably, on the ComplexMotion test set, it achieves a mask mIoU of 78.34, demonstrating effective alignment in scenarios involving complex motion.

\subsection{Case Study}
\label{sec:case_study}

\paragraph{Comparison with State-of-the-Art T2V Models.}

As previously noted, text prompts alone cannot fully describe complex actions, limiting the VBench~\cite{vbench_HuangHYZS0Z0JCW24} evaluation metrics in capturing MVideo's advantages for generating intricate movements. To address this, we provide additional visual comparisons that more intuitively highlight MVideo’s superior performance in complex action video generation. Figure~\ref{fig:vis_compare} clearly shows that, compared to text-to-video models such as OpenSora-v1.2~\cite{opensora} and CogVideoX-5b~\cite{yang2024cogvideox}, MVideo achieves significantly greater complexity, accuracy, and coherence in action representation.

\begin{figure}[!t]
    \centering
    \includegraphics[width=0.9\linewidth]{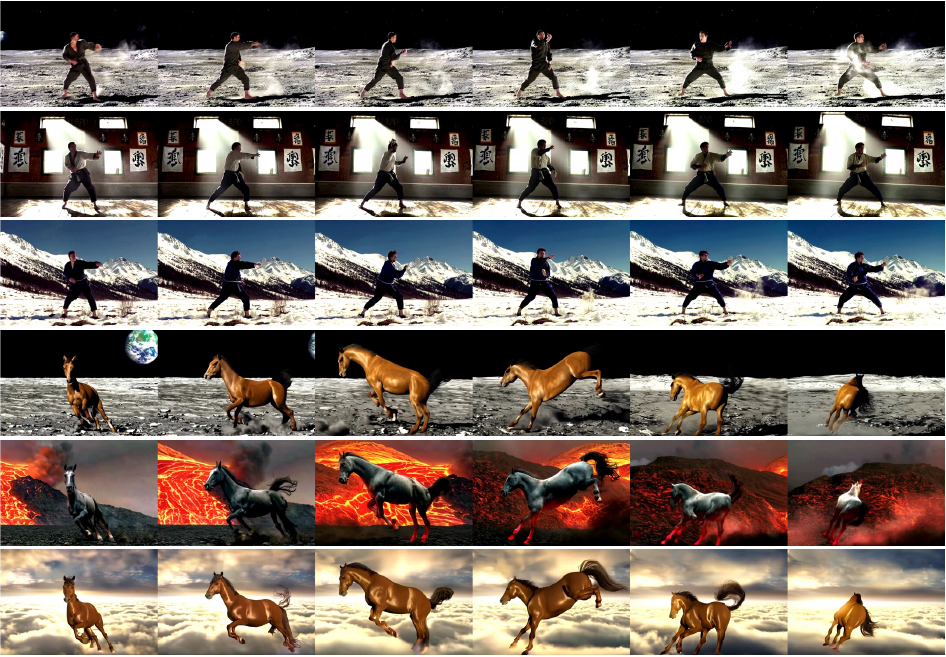}
    \caption{MVideo generates videos with identical motion conditions and varied text prompts to change scenes. Each video is 12 seconds long with a resolution of 480 $\times$ 720 pixels.}
    \label{fig:alter_text}
\end{figure}

\paragraph{Altering background scenes.}
By integrating mask sequence and text prompt alignment, MVideo enables diverse video generation. For instance, with a fixed mask sequence, MVideo can produce varied scenes by simply adjusting the text prompt, as shown in Figure~\ref{fig:alter_text}.

\paragraph{Altering moving objects.}

\begin{figure}[!t]
    \centering
    \includegraphics[width=0.9\linewidth]{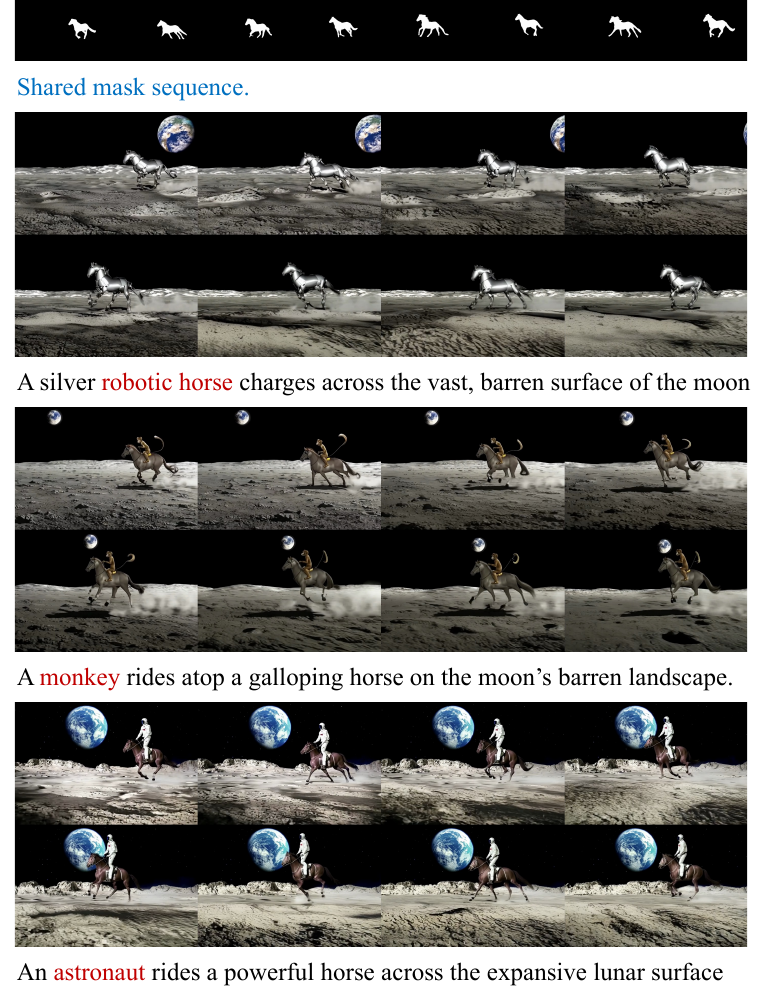}
    \caption{Videos generated with a fixed mask sequence, modifying text prompts to alter or introduce new moving subjects.}
    \label{fig:alter_foreground}
\end{figure}

Besides modifying video backgrounds via text prompts, MVideo enables altering moving objects as well. For instance, as shown in Figure~\ref{fig:alter_foreground}, given a mask sequence of a running horse, we can generate a video of either a regular horse running on the moon or a robotic horse performing the same action. Moreover, we can introduce new objects that are not in the original mask sequence. Figure~\ref{fig:alter_foreground} illustrates this capability, showing how the same horse mask sequence can be used to create videos of a monkey or an astronaut riding the horse as it runs.

\paragraph{Altering camera motion.}

\begin{figure}[!t]
    \centering
    \includegraphics[width=0.9\linewidth]{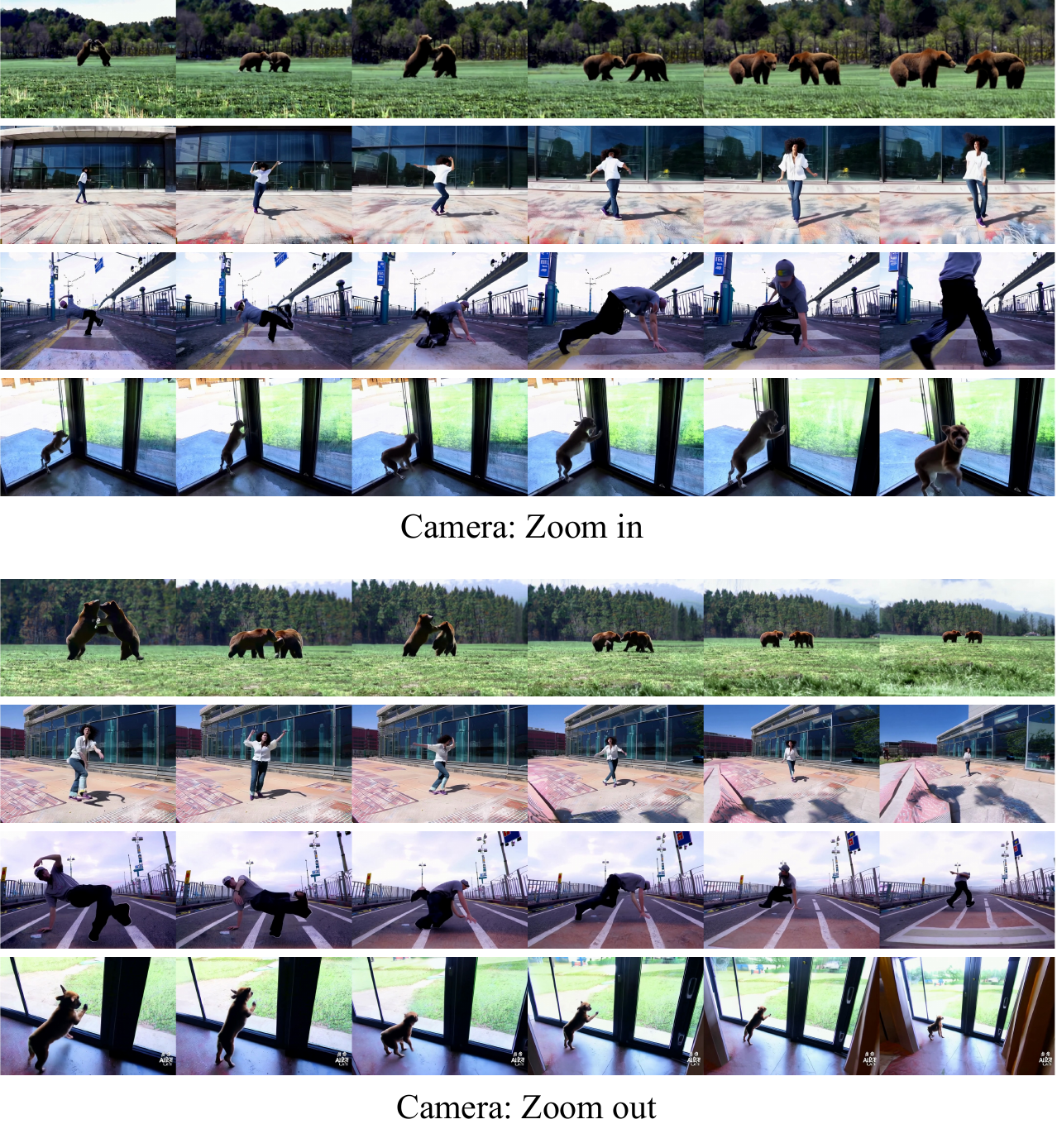}
    \caption{Different camera motion videos generated from the same mask sequence.}
    \label{fig:alter_camera}
\end{figure}

Adjusting mask scale and position enables various camera motions in videos. As shown in Figure~\ref{fig:alter_camera}, the same mask sequence produces zoom-in and zoom-out effects.

\paragraph{Editing mask sequence.}

\begin{figure}[!t]
    \centering
    \includegraphics[width=0.9\linewidth]{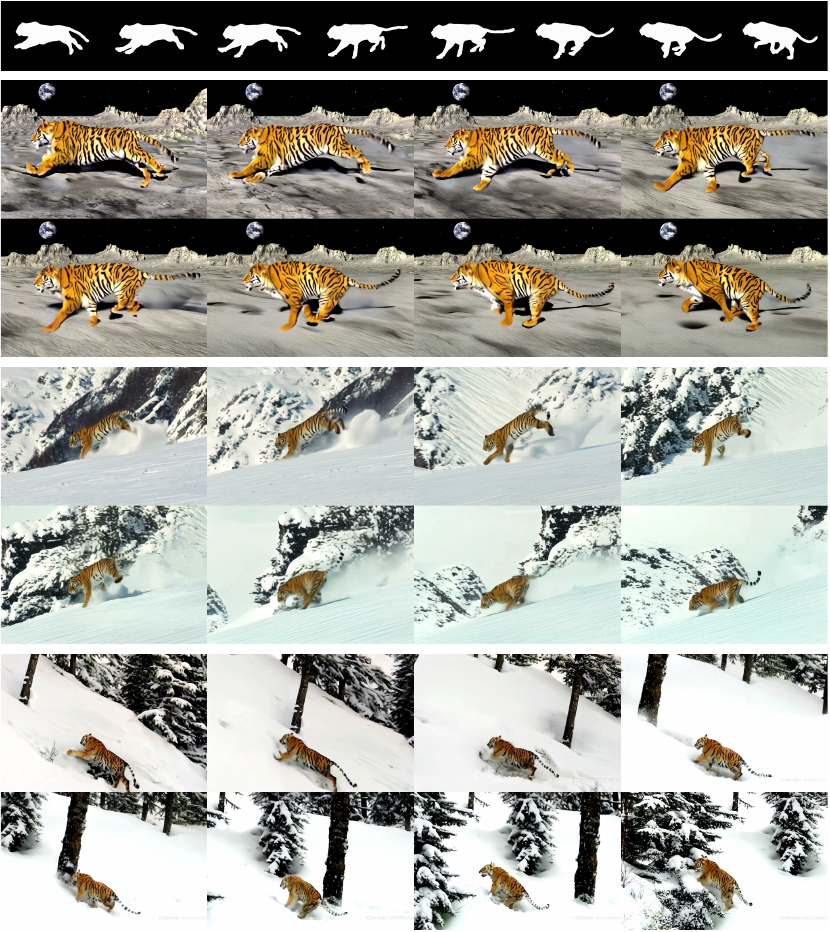}
    \caption{More dynamic videos generated by MVideo through mask sequence editing.}
    \label{fig:edit_mask}
\end{figure}

Editing an existing mask sequence allows for generating more dynamic videos. For instance, as shown in Figure~\ref{fig:edit_mask}, a mask sequence of a tiger running on flat ground can be modified to depict the tiger running uphill or downhill. This mask-editing capability significantly enhances MVideo’s versatility.

\paragraph{Combine motion conditions.}

\begin{figure}[!t]
    \centering
    \includegraphics[width=0.9\linewidth]{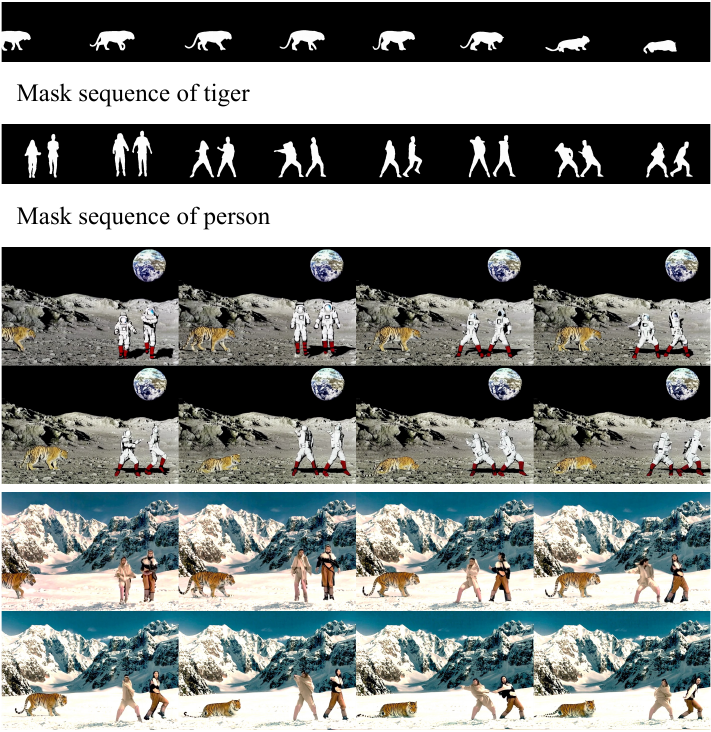}
    \caption{Videos generated by MVideo through combining multiple mask sequence conditions.}
    \label{fig:vis_motion_combination}
\end{figure}

MVideo supports the integration of multiple mask sequence conditions in video generation, enabling the creation of complex and engaging content. For example, as shown in Figure~\ref{fig:vis_motion_combination}, combining the mask sequences of a tiger and two dancing individuals allows the generated video to simultaneously capture both distinct motion patterns.

\subsection{Ablation study}

In these ablation experiments, MVideo utilizes CogVideoX-2b as the pretrained video diffusion model, with LoRA fine-tuning on mask-to-video training samples. Training is conducted over 10,000 steps on 16 GPUs with a total batch size of 32 and a learning rate of 2e-4.

\paragraph{Consistency loss.}

\begin{table}[!t]
\centering
\resizebox{\linewidth}{!}{%

\begin{tabular}{@{}c|c|c|ccc@{}}
\toprule
\multirow{3}{*}{Model} & \multirow{3}{*}{\begin{tabular}[c]{@{}c@{}}Consistency \\ loss\end{tabular}} & \multirow{3}{*}{Video} & \multicolumn{3}{c}{VBench test set}                                                                                                                                                                                                     \\ \cmidrule(l){4-6} 
                       &                                   &                        & \multicolumn{1}{c|}{\begin{tabular}[c]{@{}c@{}}overall \\ consistency\end{tabular}} & \multicolumn{1}{c|}{\begin{tabular}[c]{@{}c@{}}imaging\\ quality\end{tabular}} & \begin{tabular}[c]{@{}c@{}}motion\\ smoothness\end{tabular} \\ \midrule
    
CogvideoX-2b           &                                   &          $6s \times 480 \times 720$              & \multicolumn{1}{c|}{25.45}                                                               & \multicolumn{1}{c|}{61.48}                                                              &              98.02                                               \\ \midrule

MVideo-2b              &               &        $8s \times 480 \times 720$           & \multicolumn{1}{c|}{23.72}                                                               & \multicolumn{1}{c|}{52.67}                                                              &                 97.98                                            \\ 
MVideo-2b              &           $\checkmark$             &   $8s \times 480 \times 720$                   & \multicolumn{1}{c|}{26.48}                                                               & \multicolumn{1}{c|}{58.44}                                                              &        97.96                                                     \\ \bottomrule
\end{tabular}
}
\caption{Ablation study on consistency loss. MVideo-2b is fine-tuned from CogVideoX-2b using LoRA and is evaluated on the VBench~\cite{vbench_HuangHYZS0Z0JCW24} test set without utilizing any mask condition.}
\label{tbl:ablation_consistency_loss}
\end{table}

Our experiments show that further fine-tuning the pretrained video diffusion model on the mask-to-video dataset reduces MVideo’s text alignment performance, as indicated by declines in ``overall consistency" and ``imaging quality" metrics. As shown in Table~\ref{tbl:ablation_consistency_loss}, testing on the VBench set revealed a drop in ``overall consistency" from 25.45 to 23.72 and ``imaging quality" from 61.48 to 52.67.
To address this degradation, we introduced a consistency loss that effectively preserves MVideo’s text alignment and imaging quality. As demonstrated in Table~\ref{tbl:ablation_consistency_loss}, adding the consistency loss during training significantly improves both ``overall consistency" and ``imaging quality" metrics.

\paragraph{Long motion video.}

\begin{table}[!t]
\centering
\resizebox{0.9\linewidth}{!}{%
\begin{tabular}{@{}c|c|ccc@{}}
\toprule
\multirow{3}{*}{Model}     & \multirow{3}{*}{Video} & \multicolumn{3}{c}{VBench test set}                                                                                                                                                                                                     \\ \cmidrule(l){3-5} 
                           &                        & \multicolumn{1}{c|}{\begin{tabular}[c]{@{}c@{}}overall \\ consistency\end{tabular}} & \multicolumn{1}{c|}{\begin{tabular}[c]{@{}c@{}}imaging\\ quality\end{tabular}} & \begin{tabular}[c]{@{}c@{}}motion\\ smoothness\end{tabular} \\ \midrule
                           
\multirow{4}{*}{MVideo-2b} & $4s \times 480 \times 720$         & \multicolumn{1}{c|}{25.55}                                                               & \multicolumn{1}{c|}{58.39}                                                              &                          98.20                                   \\ 

                           & $8s \times 480 \times 720$        & \multicolumn{1}{c|}{26.48}                                                               & \multicolumn{1}{c|}{58.44}                                                              &        97.96                                                     \\ 
                           
                           & $12s \times 480 \times 720$        & \multicolumn{1}{c|}{25.71}                                                               & \multicolumn{1}{c|}{57.76}                                                              &              97.83                                               \\ 

                           & $16s \times 480 \times 720$        & \multicolumn{1}{c|}{26.14}                                                               & \multicolumn{1}{c|}{56.88}                                                              &              98.07                                              \\ 
                           
                           \bottomrule
\end{tabular}
}
\caption{Ablation on long video generation. MVideo-2b is fine-tuned from CogVideoX-2b using LoRA and is evaluated on the VBench~\cite{vbench_HuangHYZS0Z0JCW24} test set without using any mask condition. All generated videos are produced at a frame rate of 8 FPS.}
\label{tbl:ablation_long_video}
\end{table}

MVideo iteratively generates long motion videos by generating video clips of 4 seconds each, which are then concatenated to form a complete long video. To ensure temporal consistency in the generated long video, MVideo introduces high-resolution image conditions and low-resolution video conditions to enhance both content and temporal coherence. As shown in Table~\ref{tbl:ablation_long_video}, we evaluated the performance of MVideo-2b on the VBench test set by generating videos of varying lengths. The results indicate that MVideo-2b successfully scales the video duration from 4 seconds to 16 seconds while maintaining stable overall evaluation metrics, with no significant drop in performance as the length increases.

\section{Conclusion}

In this work, we introduce MVideo, a novel framework for generating complex action videos with improved precision and temporal consistency. Unlike traditional models that rely solely on text prompts, MVideo leverages mask sequences as an additional conditioning input, offering a clearer and more accurate depiction of intricate movements. This method allows for a better capture of dynamic action sequences, addressing key limitations in current video generation models. Moreover, MVideo’s iterative generation approach balances computational efficiency with temporal coherence, enabling the creation of longer, narratively cohesive videos without sacrificing consistency.
Our experiments demonstrate MVideo's strong generalization in mask alignment and its ability to generate complex action videos by editing or combining mask sequences.

{
    \small
    \bibliographystyle{ieeenat_fullname}
    \bibliography{main}
}


\end{document}